# Uneven illumination surface defects inspection based on convolutional neural network

Hao Wu
*School of Mechanical Engineering,*
*Anhui University of Technology*
Maanshan, China
wuhaomoses@qq.com

Yulong Liu
*School of Mechanical Engineering,*
*Anhui University of Technology*
Maanshan, China
dulyam@163.com

Wenbin Gao
*School of Mechanical Engineering,*
*Anhui University of Technology*
Maanshan, China
349282565@qq.com

Xiangrong Xu
*School of Mechanical Engineering,*
*Anhui University of Technology*
Maanshan, China
xuxr88@qq.com

***Abstract*—The Surface defect inspection based on machine vision is affected by uneven illumination. In order to improve the inspection rate of surface defects inspection under uneven illumination condition, this paper proposes a method for detecting surface image defects based on convolutional neural network, which is based on the adjustment of convolutional neural networks, training parameters, changing the structure of the network, to achieve the purpose of accurately identifying various defects. Experimental on defect inspection of copper strip and steel images shows that the convolutional neural network can automatically learn features without preprocessing the image, and correct identification of various types of image defects affected by uneven illumination, thus overcoming the drawbacks of traditional machine vision inspection methods under uneven illumination.**

## I. INTRODUCTION

During inspection of the product, the surface defect inspection is a very important part, the surface area of the defect is localized damage to the surface when the object is inconsistent with the peripheral portion, or deface the surface means, for example: hole, dust, scratches, abrasions, dents, etc. there are defects on the surface in the object, in the course of manufacture can cause serious problems, in addition to the economic value of the product affects its appearance is more likely to affect the internal functionality of the product, reduce. The surface defect inspection technology based on machine vision has gradually become a research hotspot recently, and there have been many researches and applications [1-3].It can replace the human eye to make more accurate measurements and judgments [4-5].It is based imaging, image processing and analysis of the automatic inspection and analysis equipment, has made the development of a plurality of field [6 ], such as the strip and electronics .

Machine vision based surface defects inspection method can be divided into the following categories [7]: statistical-based methods, structure-based methods, spectrum-based methods, and subspace-based methods, of which the latter two are used the most.

Statistical-based methods include optimal threshold method [8], gray level co-occurrence matrix method [9] and LBP (Local Binary Patterns) [10], etc., mainly by measuring the statistical characteristics of pixel spatial distribution. However, the main disadvantage of this type of method is that it performs poorly on textures composed of larger-sized primitives and requires high computational performance because a large number of adjacent pixels are calculated. This type of method cannot handle the problem of uneven gray scale caused by uneven illumination.

The structure-based method mainly uses the feature of similarity repeated texture in texture image [11], using a primitive texture as the basis, and then combining the primitives according to certain rules to obtain a defective free texture image. The difficulty is that many textures vary widely, and are not simple similar repetitions. It is impossible to adjust the algorithm to match the specific geometry of the defect, nor to the structure without regularity. The uneven illumination caused in the gray value of the image is less suitable to be matched by the structural method.

The spectrum method is transformed into the frequency domain by time-frequency conversion, mainly including Fourier transform method [12], Gabor transform method [13] and wavelet transform method [14]. The Fourier transform method is a global transform that only contains frequency domain information and does not contain time information. The short-time Fourier transform, named as Gabor transform, as a substitute for the Fourier transform, can extract image texture features and can well simulate the visual perception of the human visual system, sensitive to the edges of the image. However, when such methods are applied to texture image inspection, it is often necessary to have strong empirical parameters in order to make the extracted features meet the requirements.

The basic concept of the subspace-based approach is to model and determine the correlation between learning samples to represent more samples. After simple re-projection processing, redundant information and noise are reduced, and hidden information is Excavation. Common methods based on eigenvalues, such as Principal Component Analysis (PCA) [15], Linear Discriminant Analysis (LDA) and Independent Component Analysis (ICA) [16] Is a method widely used to estimate sample projection. The subspace transform method avoids complex feature design, and measures the similarity by using simple rules in the transform space. The simple matrix calculation makes the inspection process simple and fast, but also is affected by illumination unevenness.

Conventional defect classification methods based on the feature extraction, the sample’s consistency is required highly, and when illumination unevenness occurs, it can't cope well with the uneven illumination and noise , and it is a supervised learning method that requires a priori knowledge , and how to construct an unsupervised way for automatically learn how to construct the best features that can characterize the nature of the image through the sample learning is current research hotspot , deep learning (deep learning) [17] can automatically learn the best features by building a multi-layered network, and make the learned features have better generalization performance. Convolutional neural networks are the most representative type of deep learning applications in the field of image recognition. Therefore, we propose a method for automatic identification of surface defects based on convolutional neural networks.

## II. THE PROPOSED METHOD OF CONVOLUTIONAL NEURAL NETWORK STRUCTURE

Convolutional neural network is a feed forward artificial neural network, which has become a hotspot in the field of speech analysis and image recognition. Its weight share structure is more similar to biological neural network, it reduces the complexity of the network model, and reduce the number of weights. when the network input is multi-dimensional images, the advantages are more obvious. Images can be directly as a network input, avoiding the traditional recognition algorithm in complex feature extraction and data reconstruction processes. Convolutional neural networks are multi-layered sensors that are specifically designed to recognize two-dimensional shapes that are highly invariant to translation, scaling, and rotation.

LeNet-5 is one of the convolutional neural networks [18] , and its structure is shown in Figure 1 : LeNet-5 structure were seven layer structure. First as the input layer of input, then C1 is the convolution layer, which contains a Four feature map, the convolution kernel size is $na \times na$ , and the feature map size after the convolution operation is $m2 \times m2$ . S2 Sampling is the next layer, comprising feature image with number of a , each feature in C1 neurons corresponding features of $nb \times nb$ local neighborhood as input. C3 is also a convolution layer, a feature maps has the size of $m4 \times m4$ . S4 is the downsampling layer, including b feature maps, each of the neurons in each feature map of S4 corresponds to C3 feature map. Subsequent to S4 is a fully connected layer containing neurons. The final output layer consists of radial basis functions used to identify defects.

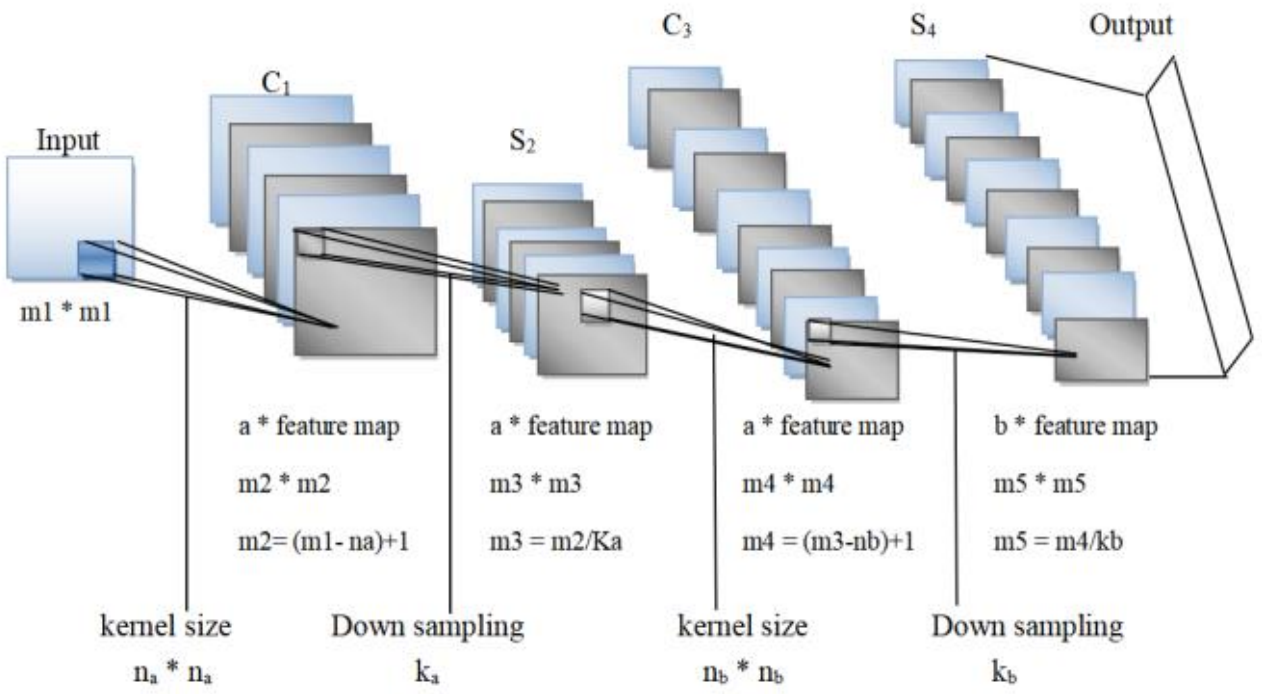

Fig. 1. Convolutional neural network structure .

## III. EXPERIMENTAL PROCESS AND RESULTS

### *A. Training process*

Firstly, collect a set of surface images such as copper strip image image samples, including 30 training images, where the normal number is 15, the defective number is 15. And 30 test images, 15 of which are normal and others defective. For simple defects such as inclusions, pits, and flaking, the defect images are shown in Figure 2 : where these defects are obvious, the right part area is darker than the left area , that is, they are affected by uneven illumination .

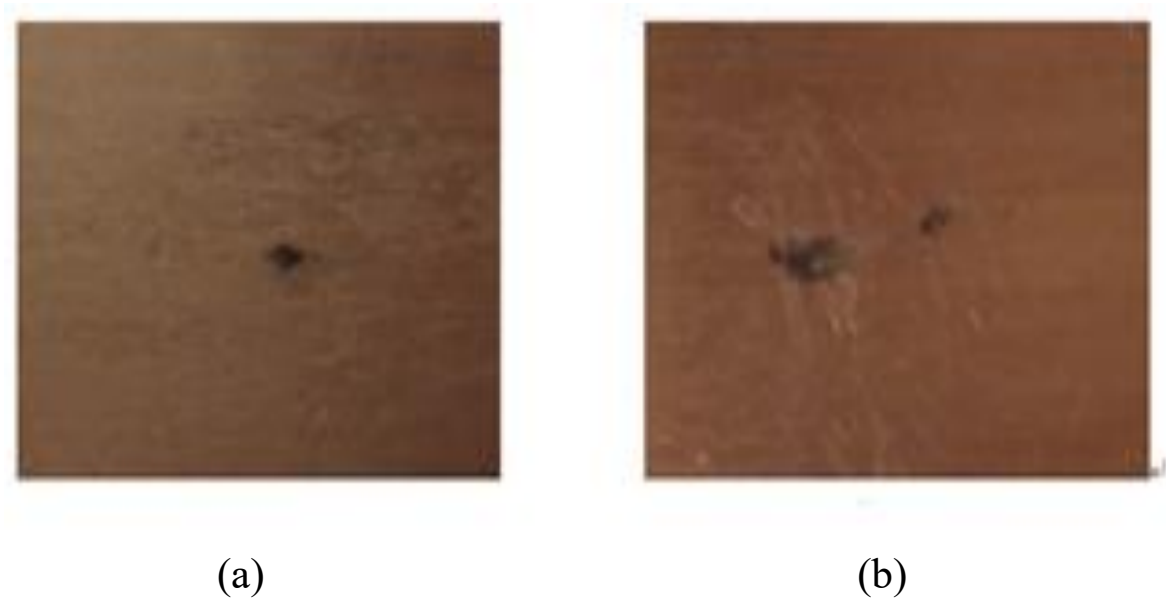

Fig. 2. Defect copper image samples. (a): inclusions; (b): pits.

The experiment platform is run on MacBook air notebook with 8GB of memory computer, this paper build the convolutional neural network, which is similar to convolutional neural network LeNet-5 , this convolutional neural networks has two convolution layers and two pooled Layer. And it is trained in the number of different numbers. Then test it.

The input image height and width are $400 \times 400$ (ie, image pixel 400 400 ), and the first layer of convolution kernel is $3 \times 3$ , convolution layer c1 , get 6 feature maps of size $398 \times 398$ , the downsampling coefficient is 2 , the downsampling layer S2 gets 6 sizes of $199 \times$ Feature map of 199 , the second layer of convolution kernel size is $3 \times 3$ , convolutional layer C3 gets 12 sizes of $100 \times 100$ feature map, the downsampling coefficient is 2 , the downsampling layer S4 , get 12 sizes of $50 \times 50$ feature map. The output layer is 2 due to classification and defect .

Because the size of the training copper images is inconsistent, there are different big and small size samples, the pixels are about $800 \times 900$ , and the images are input into the convolutional neural network model and learn and train. The first step is to enter the image into the model , so the first step is to input the image. Input the image directly using the function. Name the non-defective copper strip image " OK " (that is, pass), and make a label with its label set to 1 and the defective copper strip image image " bad " (ie defect). Make the label set its label to 0 . Obviously, our experiment is a two-category, and the experimental output only needs to classify the copper image recognition as " 1" or " 0 ". The second step is to build a convolutional neural network. The convolutional neural network we built is two convolutional layers and two pooling layers, which are relatively simple. The third step calls the first two steps, then trains the CNN , and saves the training results in the relevant folder, which can be directly applied after the test.

Since the two-layer convolution is relatively simple, we set the number of training steps to 5000 for the first time .

Based on the computer configuration, and there is no GPU built , the training is cost for 2 hours, and the correct rate and loss during the training. The function changes are shown in Figure 3 and Figure 4 below. The orange and blue curves represent changes in the training set and the validation set, respectively.

It can be clearly seen from Fig. 3 and Fig. 4 that at the beginning of training, the loss function is about 0.6 and the correct rate is about 0.5 . As the number of training steps increases, the loss function gradually decreases, and the correct rate gradually increases. When it reaches 2000 , the loss function is about 0.5 , and the correct rate is about 0.8 . When the number of training steps reaches 5000 , the loss function has been reduced to about 0 , and the correct rate is close to 1 . Therefore, according to the displayed results, the model for classifying the copper strip image defect inspection and identification is satisfactory . Therefore, the trained model is saved, and then the selected model is tested on the trained model.

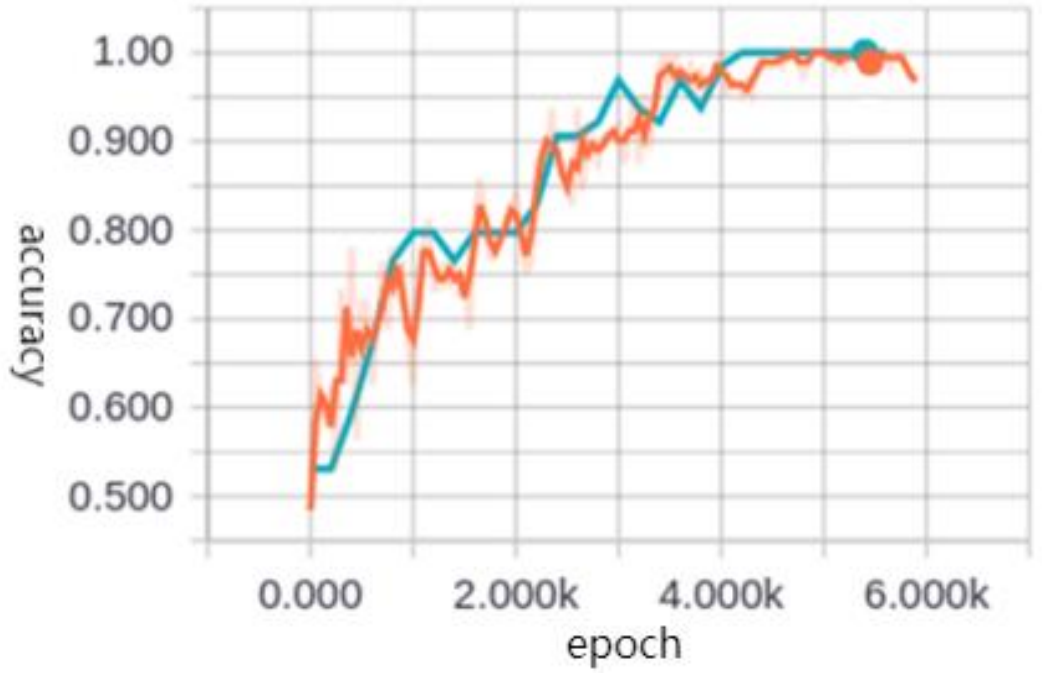

Fig. 3. Correct rate change curve: training set ( orange ) and verification set ( blue )

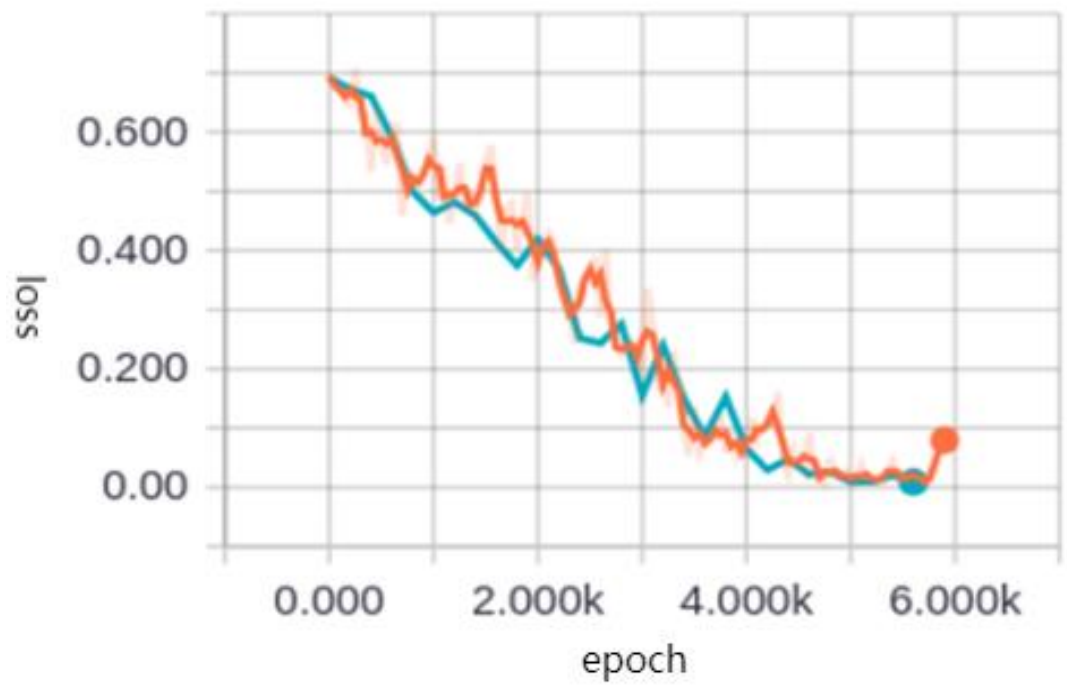

Fig. 4. Loss function curve: training set (orange) and verification set ( blue )

## B. *Test process*

The training of the convolutional neural network has been completed above , and the training results are saved. The trained model will be directly tested .

Prepared 15 defective and 15 non-defective copper bar images for testing in advance. Apply relevant read and write functions to read the trained model and test it. Here input a image each time to test. The test results of some samples are as follows:

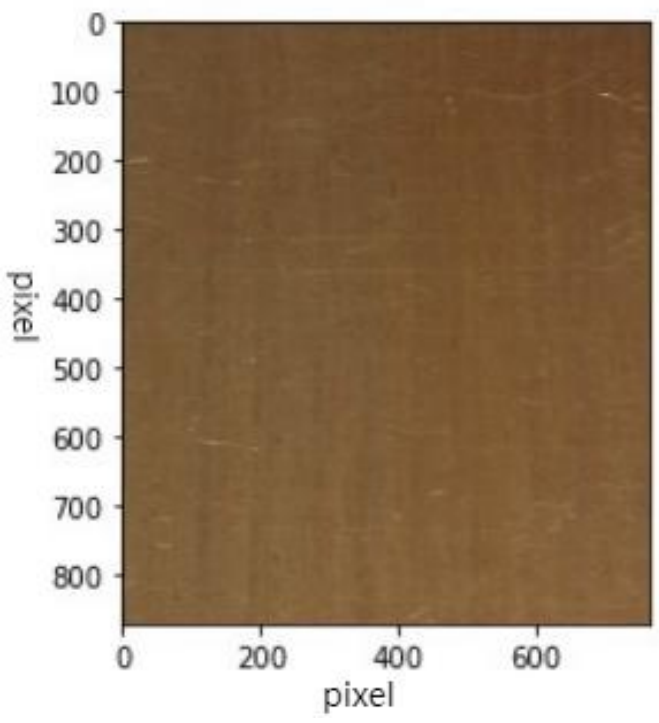

Fig. 5. Qualified sample test results

It can be seen from Figure 5 that there is no defect in the image of the copper strip image tested, it is a complete good copper strip image, and the result of the experimental model output " this is OK with possibility 0.67436 ", it is also a correct determined for the copper strip image . The probability of judgment is approximately 0.67 4 . The judgment is correct.

Furthermore, there are obvious defects in Figure 6. The test results show that the possibility of the copper strip image being defective is 0.678 . Thus, the judgment result is correct. After several consecutive tests, the experimental results were the same as above, and no judgment errors occurred.

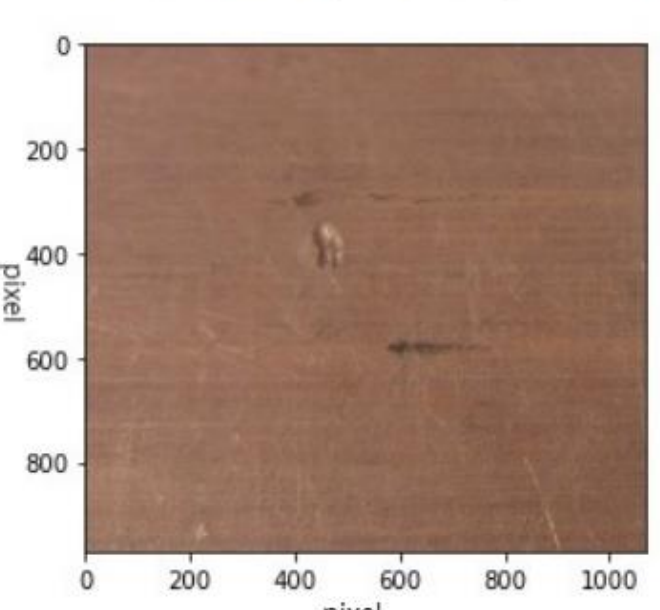

Fig. 6. Defective sample test result

# IV. Discussion and Analysis

According to the experimental results obtained above , we carry out corresponding analysis. It can be seen from the test images that the defect characteristics of the images to be tested are still obvious, but the discrimination rate is only about 0.6 to 0.7 . Consider to Improve it by:

( 1 ) In the above experiment, we have 5000 steps for the model . Consider the model structure to train more steps to improve the recognition rate.

( 2 ) The convolutional neural network model we used above is too simple, so the identification of defects is not particularly obvious. Consider building a deeper convolutional neural network model to improve recognition accuracy.

## A. *Increase the number of training steps*

For the above analysis, we first improve the number of training steps and perform 10,000 steps of training. The experimental steps are the same as above, and so it will not be described detail here again. We directly tested the defect samples . The test results are as follows: The probability of judgment is 0.764 . Compared with the above test, the recognition rate of the model is improved . In order to verify the reliability of the conclusion, we test the qualified samples again . The test results are as follows: The probability of judgment is 0. 774 . Compared with the discriminating rate of 0.67 in Figure 5 and Figure 6 , there is a higher improvement. Therefore, it can be concluded that the number of training steps has a certain influence on the identification of the model we built. Next, we performed 15000 steps and 20000 steps of training. After the training, the same test step was performed. The test results are shown in Table 1 : As the number of training steps increases, the model has better training, so the defects are discriminated. At the time, the discriminating rate has also been significantly improved. Therefore, it can be seen that as the number of training steps increases, the model discriminating rate rises steadily.

TABLE I. THE INFLUENCE OF THE NUMBER OF TRAINING STEPS ON THE RECOGNITION ACCURACY

| Number of training steps | qualified | defect |
| --- | --- | --- |
| 5000 | 0. 674 | 0.678 |
| 10000 | 0. 764 | 0.774 |
| 15000 | 0 . 874 | 0.884 |
| 20000 | 0.987 | 0.986 |

### *B. Change convolutional neural network structure*

Next, we also consider the use of a deeper convolutional neural networks, build three convolutional layers, three pooled layer of convolutional neural networks, perform the same operation, because the use of deeper network structure will increase training time, So here, only 5000 steps of training are performed , and the same test step is performed at the end of the training, and the results of the deep convolutional neural network for the image will be compared . When applying a three- layer convolutional neural network, the discriminant rate obtained by the experiment is 0.98 . Compared with the convolution of the two layers, the number of training steps is 5000 , which improves the recognition rate of about 0.3 , which is significantly improved. All the images were tested later, and the model showed no errors, and the discriminant rate was constant at around 0.97.

This experiment finally determined the application of convolutional neural network ( CNN ) model to identify copper strip image defects, and made appropriate comparisons for different training steps. The corresponding conclusion is obtained: the number of training steps of the convolutional neural network model will affect its judgment. Under certain conditions, the more training steps, the more accurate the judgment. In addition, on the basis of simple CNN , a deeper convolutional neural network structure is built, and the results show that the recognition accuracy is significantly improved.

### *C. Experimental Validation on NEU surface defect dataset*

In order to further verify our proposed method, the NEU surface defect database[19] is used, this dataset contain several types of steel defect, here two typical surface defects of hot-rolled steel strips are used for our proposed method in this paper, i.e. patches (Pa), scratches (Sc). There are 300 samples for each type of defect, we use 250 samples for training(80%) and validation(randomly choose 20% of the whole 250 samples as validation data set), and 50 samples for testing, when the training epoch is 7, the training accuracy is over 90%, and the training accuracy arrived 100% when the training epoch is 13;when the training epoch is 44,the validation accuracy is 100%; when the training epoch is 146, the validation loss is 0.001,after 220 epoch of training, the validation loss is zero.

For testing results, the inspection accuracy of patches is varied from 99.0539% to 99.9998%, the average accuracy for all the testing set of patches defect is about 99.8805%. The inspection accuracy of scratches is varied from 98.2197% to 99.99999%, the average accuracy for all the testing set of scratches is 99.7739%.

## V. CONCLUSION

In this paper, experimental results is obtained through inspection methods based on convolutional neural network to identify defects on the surface, as following:

( 1 ) The essence of the convolutional neural network is to let the computer learn autonomously in input data. After obtaining the "experience", it makes judgments on the same object , and no longer requires artificially specified rules . In general, it is similar to the process of learning new knowledge by human. Compared with the simple neural network structure, the deep convolutional network structure has more powerful learning ability and is better at extracting complex features . This is why the discriminant rate increases significantly when a layer of convolution is added to the experiment . At the same time, the time required for the deep structure to train will increase significantly .

( 2 ) The proposed method in this paper can detect defects on the surface of copper strip images under the influence of uneven illumination , it overcomes the limitations of traditional machine vision inspection methods, and adjust the structure of convolutional neural network to change the structure of the network and improve the identification rate .

( 3 ) The images used in the tests in this paper are generally common defects such as scratches and pits . It is worthwhile to conduct further research for more defective samples, especially those with difficult to recognition for other traditional visual methods and human eyes.

## ACKNOWLEDGMENT

This research was supported in part by the China National Key Research and Development project (2017YFE0113200), National Natural Science Foundation of China ( No.51605004) ,Anhui Provincial Natural Science Foundation(1808085QE162 ; 2108085ME166), Natural Science Research Project of Universities in Anhui Province(KJ2021A0408) ,Open Project of Anhui Province Key Laboratory of Special and Heavy Load Robot (TZJQR007-2021).The authors also would like to thank Dr Jan Paul Siebert, School of Computing Science University of Glasgow, for his friendly host and instruction during the author visiting research in UK. This support is greatly acknowledged.

## References

[1] Zhang Jing , Ye Yutang ,& Xie Wei (2014). Photoelectric inspection of defects in metal cylindrical workpieces [J] .Optical Precision Engineering , 22(7): 1871-1876.

[2] Chu, Weishen, et al. "Effects of Wiring Density and Pillar Structure on Chip Package Interaction for Advanced Cu Low-k Chips." 2020 IEEE International Reliability Physics Symposium (IRPS). IEEE, 202..

[3] Chu, Weishen, Tengfei Jiang, and Paul S. Ho. "Effect of Wiring Density and Pillar Structure on Chip Packaging Interaction for Mixed-Signal Cu Low k Chips." IEEE Transactions on Device and Materials Reliability (2021).

[4] Wang, F. L., &Zuo, B. (2016). Inspection of surface cutting defect on magnet using Fourier image reconstruction. Journal of Central South University, 23(5), 1123-1131.

[5] Xuezhi, Y., Haiqin, Z., Yuan, C., Kewei, W., & Zhao, X. (2013). Fabric defect inspection based on PCA-NLM. Journal of Image and Graphics, (12), 4.

[6] Yishu, S. K. Y. Y. P., &Dewei, D. O. N. G. (2012). Convex active contour segmentation model of strip steel defects image based on local information. Journal of Mechanical Engineering, 48(20), 20.

[7] Xie, X. (2008). A review of recent advances in surface defect inspection using texture analysis techniques. ELCVIA: electronic letters on computer vision and image analysis, 7(3), 1-22.

[8] Truong, M. T. N., & Kim, S. (2018). Automatic image thresholding using Otsu's method and entropy weighting scheme for surface defect inspection. Soft Computing, 22(13), 4197-4203.

[9] Chondronasios, A., Popov, I., & Jordanov, I. (2016). Feature selection for surface defect classification of extruded aluminum profiles. The International Journal of Advanced Manufacturing Technology, 83(1-4), 33-41.

[10] Ko, J., &Rheem, J. (2016). Defect inspection of polycrystalline solar wafers using local binary mean. The International Journal of Advanced Manufacturing Technology, 82(9-12), 1753-1764.

[11] Soille, P. (2013). Morphological image analysis: principles and applications. Springer Science & Business Media..

[12] Tsai, D. M., Wu, S. C., & Li, W. C. (2012). Defect inspection of solar cells in electroluminescence images using Fourier image reconstruction. Solar Energy Materials and Solar Cells, 99, 250-262.

[13] Tong, L., Wong, W. K., &Kwong, C. K. (2016). Differential evolution-based optimal Gabor filter model for fabric inspection. Neurocomputing, 173, 1386-1401.

[14] Jeon, Y. J., Choi, D. C., Lee, S. J., Yun, J. P., & Kim, S. W. (2014). Defect inspection for corner cracks in steel billets using a wavelet reconstruction method. JOSA A, 31(2), 227-237.

[15] Liu, H. W., Chen, S. H., &Perng, D. B. (2016). Defect inspection of patterned thin-film ceramic light-emitting diode substrate using a fast randomized principal component analysis. IEEE Transactions on Semiconductor Manufacturing, 29(3), 248-256.

[16] Tsai, D. M., Wu, S. C., & Chiu, W. Y. (2013). Defect inspection in solar modules using ICA basis images. IEEE Transactions on Industrial Informatics, 9(1), 122-131.

[17] Hinton, G. E., Osindero, S., &Teh, Y. W. (2006). A fast learning algorithm for deep belief nets. Neuralcomputation, 18(7), 1527-1554.

[18] LeCun, Y., Bottou, L., Bengio, Y., &Haffner, P. (1998). Gradient-based learning applied to document recognition. Proceedings of the IEEE, 86(11), 2278-2324.

[19] Song, K., & Yan, Y. (2013). A noise robust method based on completed local binary patterns for hot-rolled steel strip surface defects. Applied Surface Science, 285, 858-864.